\documentclass[runningheads]{llncs}
\usepackage[T1]{fontenc}
\usepackage{graphicx}
\usepackage[linesnumbered, ruled]{algorithm2e}
\usepackage{amsmath,amsfonts}
\usepackage{booktabs}
\usepackage{tikz}
\usepackage{gensymb}
\usetikzlibrary{shapes.geometric, arrows}
\usepackage{multirow}
\usepackage{float}
\begin{document}
\title{Probabilistic Verification of Neural Networks via Efficient Probabilistic Hull Generation}

\titlerunning{Probabilistic Verification of Neural Networks}
%
\author{Jingyang Li\inst{1}
\and
Xin Chen \inst{2}
\and
Hongfei Fu\inst{1}
\and
Guoqiang Li\inst{1}
}
\authorrunning{J. Li et al.}
\institute{Shanghai Jiao Tong University, Shanghai 200240, China\\
\email{\{lijjjjj, jt002845, li.g\}@sjtu.edu.cn}
\and
University of New Mexico, Albuquerque, NM 87106, USA\\
\email{chenxin@unm.edu}\\
}
\maketitle              
\begin{abstract}
The problem of probabilistic verification of a neural network investigates the probability of satisfying the safe constraints in the output space when the input is given by a probability distribution. It is significant to answer this problem when the input is affected by disturbances often modeled by probabilistic variables. In the paper, we propose a novel neural network probabilistic verification framework which computes a guaranteed range for the safe probability by efficiently finding safe and unsafe probabilistic hulls. Our approach consists of three main innovations:
(1) a state space subdivision strategy using regression trees to produce probabilistic hulls, (2) a boundary-aware sampling method which identifies the safety boundary in the input space using samples that are later used for building regression trees, and (3) iterative refinement with probabilistic prioritization for computing a guaranteed range for the safe probability. The accuracy and efficiency of our approach are evaluated on various benchmarks including ACAS Xu and a rocket lander controller. The result shows an obvious advantage over the state of the art.

\end{abstract}
\section{Introduction}
Formal verification of deep neural networks (DNNs) is crucial for ensuring the trustworthiness and the reliability of learning-enabled systems in safety-critical scenarios~\cite{DBLP:journals/corr/abs-2206-12227,DBLP:journals/ase/TambonLANMPKAML22}. While most existing techniques aim at solving the output range analysis problem of DNNs~\cite{Katz2017ReluplexAE,DBLP:conf/cav/WuIZTDKRAJBHLWZKKB24,Zhang2018EfficientNN}, it is still difficult to know the safety of a DNN when the input is affected by noises that are often modeled by random variables. Since noises or disturbances are often described by Gaussian variables, their values are not bounded and can hardly be handled by existing output range analysis methods. We propose a new approach which generates a guaranteed range for the safety of a DNN when its input is Gaussian variables.





Probabilistic verification~\cite{PROVEN,ProbabilisticPracticalRobustness,StatisticalRobustnessNN,PVBranchBound,QuantitativeVerificationProbStars} involves analyzing the probability of safety violations under specified input distributions instead of giving a deterministic answer to the safety query. In order to estimate the probability of satisfying the safe constraints in the output space of a DNN, a straightforward approach could be to compute the output distribution and then integrate its density function over the safe output space. However, it is often not feasible due to the complexity of DNNs. In this paper, we propose a method which conservatively estimate the safety of a DNN based on a set of safe or unsafe \emph{probabilistic hulls} that at most intersect on facets. A probabilistic hull is a closed and bounded set in the input space whose orientation is aligned with the Gaussian distribution of the input, such that the probability of it can be efficiently computed based on the error function~\cite{Bain92}. An upper bound of safety can be obtained by subtracting the probabilities of all probabilistic hulls which are purely unsafe, while a lower bound can be derived by adding the probabilities of all safe hulls. The main contribution of our method is an efficient search strategy for finding large safe and unsafe hulls.



Existing probabilistic neural network verification methods suffer from the curse of dimensionality. Typically, the existing set-based techniques~\cite{Zhang2018EfficientNN,Wang2021BetaCROWNEB} often prioritize partitioning the input space uniformly that cannot recognize critical regions. This leads to exponential amount of subdivisions and is time consuming. Other methods, such as ProbStar~\cite{QuantitativeVerificationProbStars} symbolically propagates the input set to the output space and then estimate their safe probability, it is able to more accurately track the input-output dependency however cannot handle the DNNs whose activations are not ReLU.




We study the probabilistic verification problem on feedforward DNNs. 
We propose a novel regression tree-guided probabilistic verification framework that achieves both efficiency and generality. Our key insight is that neural network decision/safe boundaries, i.e., the boundary dividing the safe and unsafe set in the output space, can be effectively approximated through adaptive region partitioning guided by sampling-based regression trees, eliminating the unnecessary exploration and subdivision in the regions that are purely safe or unsafe.
Our approach makes three primary contributions: (1) We introduce a regression tree-guided state space splitting strategy for the input space that efficiently generates purely safe or unsafe probabilistic hulls of large probabilities. It avoids the unnecessary subdivisions in the safe or unsafe regions in most practical cases. (2) We propose a boundary-aware sampling method which effectively identifies the safety boundary in the input space by random samples which are later used to build regression trees. (3) We develop an incremental sampling mechanism that iteratively refines the undecided regions with the largest probability, avoiding the prohibitive cost of dense sampling across the entire input space.
Experimental evaluation on the challenging benchmarks demonstrates that
our method achieves superior accuracy against the state of the art and achieves up to 10× speedup compared to basic branch-and-bound approaches.


\section{Related Work}



Linear bound propagation techniques, exemplified by CROWN~\cite{Zhang2018EfficientNN}, have revolutionized neural network verification by extending traditional interval bound propagation to linear function-based bound propagation. This advancement significantly improves scalability while maintaining reasonable tightness for verification tasks. Building upon CROWN, $\beta$-CROWN~\cite{Wang2021BetaCROWNEB} integrates branch-and-bound (BaB) techniques with GPU acceleration to achieve both completeness and scalability in verification processes.


Probabilistic verification has emerged as a critical paradigm for quantifying neural network safety under stochastic inputs.
One part of the research focus on Bayesian neural network verification~\cite{DBLP:conf/ijcai/CardelliKLPPW19,DBLP:journals/ai/WickerLPPAK24,DBLP:conf/fm/WickerPLK24}.
Early approaches like PROVEN~\cite{PROVEN} established distribution-aware robustness guarantees by integrating probabilistic measures with formal verification techniques. Subsequently, hybrid methods~\cite{ProbabilisticPracticalRobustness} combined abstract interpretation with Monte Carlo sampling to balance computational efficiency with probabilistic assurances. Statistical evaluation frameworks~\cite{StatisticalRobustnessNN} further advanced the field by employing adaptive sampling strategies to estimate perturbation tolerance while quantifying violation probabilities. Besides, ~\cite{DBLP:conf/l4dc/PilipovskySOT23} proposes a characteristic-function-based probabilistic verification method which estimates the safe probability via frequency-domain.

Recent developments have focused on scalability. The probabilistic branch-and-bound approach~\cite{PVBranchBound} iteratively refines probability bounds through dynamic region subdivision, achieving improved verification success rates but suffering from exponential complexity in high-dimensional spaces. Most notably, ProbStar~\cite{QuantitativeVerificationProbStars} introduces probabilistic star-based reachability analysis for efficient uncertainty propagation in safety-critical systems. However, ProbStar is limited to ReLU networks due to its reliance on linear region enumeration, restricting its applicability to networks with generic activation functions.


\section{Preliminaries}

A deterministic input to a DNN can be denoted by a column vector $\mathbf{x} = (x_1,\dots,x_d)$. Therefore, a noise-affected input can be denoted by $\mathbf{x} + \mathbf{w}$ where $\mathbf{w} = (w_1,\dots,w_d)$ are the noises for each input variable. In this paper, we focus on Gaussian noises, i.e., $\mathbf{w}$ subject to a multivariate Gaussian distribution, since they are the most popular models for noises in practice. Since Gaussian distributions are closed under affine transformations, a noise-affected input can be represented by Gaussian variables, i.e., $\mathbf{x} \sim \mathcal{N}(\mu, \Sigma)$ where $\mu$ is the mean and $\Sigma$ is the covariance matrix.

\paragraph{Problem Formulation.}
Let $f: \mathbb{R}^d \to \mathbb{R}^m$ denote the input-output mapping of a (feedforward) DNN which has $d$ inputs and $m$ outputs. Given a Gaussian input $\mathbf{x} \sim \mathcal{N}(\mu, \Sigma)$ and a safety property in the output space: $\Gamma(\mathbf{y}): \mathbf{C} \mathbf{y} \geq \mathbf{a}$, such that the property is defined by a finite set of affine constraints/inequalities. Here, $\mathbf{y}$ is a point/vector in the output space, i.e., $\mathbf{y} = f(\mathbf{x})$, $\mathbf{C} \in \mathbb{R}^{k \times m}$ is a matrix of coefficients, and $\mathbf{a} \in \mathbb{R}^k$ is a constant vector. The \emph{DNN probabilistic verification problem} asks to estimate the probability
\begin{equation}\label{eq:safe_prob}
 \mathbb{P}(\Gamma(\mathbf{y})\,|\,\mathbf{x} \sim \mathcal{N}(\mu, \Sigma)).
\end{equation}

As we pointed out that an exact calculation of the probability (\ref{eq:safe_prob}) is intractable. Therefore, we conservatively estimate an upper bound as well as a lower bound for the actual probability via computing a finite set of safe and unsafe probabilistic hulls.

Given a set of Gaussian variables $\mathbf{x} \sim \mathcal{N}(\mu, \Sigma)$ such that $\Sigma = diag(\sigma_1^2,\dots,\sigma_d^2)$, i.e., all variables subject to independent distributions. The probability $\mathbb{P}(\mathbf{x} \in H)$, or simply $\mathbb{P}(H)$, where the hull $H$ is a box $[a_1,b_1]\times\cdots \times [a_d,b_d]$ can be computed by evaluating the expression $\prod_{1\leq i\leq d} \mathbb{P}(a_i\leq x_i \leq b_i)$ such that
\[
 \mathbb{P}(a_i\leq x_i \leq b_i) = \frac{1}{2}\left(\text{erf}(\frac{b_i - \mu}{\sqrt{2}\sigma_i}) - \text{erf}(\frac{a_i - \mu}{\sqrt{2}\sigma_i})\right)
\]
where $\text{erf}$ is the error function~\cite{Bain92}. For any full rank covariance matrix $\Sigma$, a hull $H$ aligned with its orientation can be calculated in a similar way. That is, we may transform $\Sigma$ to a diagonal matrix, transform $H$ to a box using the same mapping and evaluate the probability of being in $H$ with the transformed distribution. Hence, for simple denotation, we assume that $\Sigma$ is diagonal.
In a DNN verification problem, a hull $H$ is safe (unsafe resp.) when all points in its output set $f(H)$ satisfying (violating resp.) the safe property $\Gamma$.

\begin{figure}[htbp]
\centering
 \includegraphics[width=0.96\linewidth]{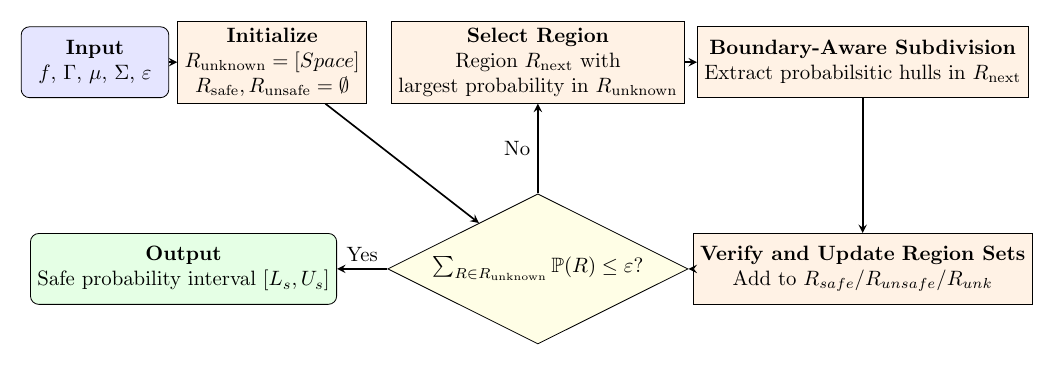}
\caption{Workflow of our regression tree-guided probabilistic verification framework. }
\label{fig:workflow}
\end{figure}

\section{Main Approach}

The main idea of our approach for estimating conservative upper and lower bounds for the actual safe probability is to iteratively construct a set of safe hulls: $H_1^s,\dots, H_N^s$ and a set of unsafe hulls: $H_1^u,\dots, H_M^u$, such that all hulls intersect at most on the facets. Therefore, each hull represents an independent probability from the others. Then, the upper bound is computed as $U_s = 1 - \sum_{1\leq i\leq M}\mathbb{P}(H_i^u)$, while the lower bound is obtained as $L_s = \sum_{1\leq i\leq N}\mathbb{P}(H_i^s)$.

\begin{theorem}
 If $H_1^s,\dots, H_N^s$ are safe and $H_1^u,\dots, H_M^u$ are unsafe, then we have that $U_s \geq L_s$, and the actual safe probability must be in the interval $[L_s,U_s]$.
\end{theorem}

Obviously, if the probabilistic hulls can be efficiently found and have high probabilities, an accurate range for safe probability can be obtained.
Unlike the related methods, the probabilistic hulls are not generated from random samples. Instead, we build (safe) boundary-ware regression trees which subdivide the input space into box regions, then we use CROWN to verify the safety of them and identify the safe and unsafe subregions as hulls. In the experiments, we show that such a strategy greatly outperforms the method using uniform sampling.






The high-level workflow of our framework is given in Figure~\ref{fig:workflow}. Given a DNN $f$, a safety property $\Gamma$, a Gaussian distribution $\mathcal{N}(\mu, \Sigma)$ for the input, and a termination threshold $\varepsilon > 0$. Here, we only need to use the DNN as a blackbox. The main algorithm keeps three sets of probabilistic hulls: $R_{\text{safe}},R_{\text{unsafe}}, R_{\text{unknown}}$ that are used to keep the set of safe, unsafe and unknown hulls respectively. Initially, $R_{\text{unknown}}$ has a single element which represents the whole state space, while $R_{\text{safe}}$, $R_{\text{unsafe}}$ are both empty. The whole state space can be represented by a large bounded set which is of a probability sufficiently close to $1$ (e.g. $\geq 99.999\%$) for the inputs, instead of an unbounded support range from Gaussian distributions.
Then the main loop performs the following steps: (1) It chooses the hull $R_{\text{next}}$ which has the highest probability in $R_{\text{unknown}}$ and delete it from $R_{\text{unknown}}$; (2) $R_{\text{next}}$ is subdivided using our boundary-aware subdivision strategy which builds a regression tree and the subdivisions are new probabilistic hulls obtained as the tree leafs; (3) We verify the output sets of the subdivisions. We add the safe hulls to $R_{\text{safe}}$, the unsafe hulls to $R_{\text{unsafe}}$, and the hulls that are not purely safe or unsafe to $R_{\text{unknown}}$. The loop terminates when the summation of the probabilities of the hulls in $R_{\text{unknown}}$ is $<\varepsilon$.
The details are described in the following subsections.

\subsection{Boundary-Aware Subdivision Strategy}


\begin{figure}[htbp]
\centering
\begin{tikzpicture}[scale=0.7,
    declare function={
        boundary(\x) = 0.8 + 0.6*sin(120*\x) + 0.4*(\x-2);
    }
]

\begin{scope}
    \fill[gray!10] (0,0) rectangle (4,3);
    \draw[thick] (0,0) rectangle (4,3);

    \draw[very thick, red, domain=0:4, samples=100] plot (\x, {boundary(\x)});

    \draw[thick, blue, dashed] (2,0) -- (2,3);
    \draw[thick, blue, dashed] (0,1.5) -- (4,1.5);

    \fill[blue] (0.8,0.7) circle (2pt);
    \fill[blue] (0.8,2.3) circle (2pt);
    \fill[blue] (2.0,0.7) circle (2pt);
    \fill[blue] (2.0,2.3) circle (2pt);
    \fill[blue] (3.2,0.7) circle (2pt);
    \fill[blue] (3.2,2.3) circle (2pt);

    \node[font=\small] at (2,3.3) {Uniform Subdivision};

    \node[orange] at (1,0.75) {\scriptsize Unk};
    \node[orange] at (3,0.75) {\scriptsize Unk};
    \node[green] at (1,2.25) {\scriptsize Safe};
    \node[orange] at (3,2.25) {\scriptsize Unk};

\end{scope}

\begin{scope}[xshift=5cm]
    \fill[gray!10] (0,0) rectangle (4,3);
    \draw[thick] (0,0) rectangle (4,3);

    \draw[very thick, red, domain=0:4, samples=100] plot (\x, {boundary(\x)});

    \draw[thick, blue, dashed] (0,1.0) -- (4.0,1.0);
    \draw[thick, blue, dashed] (2.7,1.0) -- (2.7,3);
    \draw[thick, blue, dashed] (2.9,1.0) -- (2.9,0);

    \fill[blue] (0.5, {boundary(0.5)+0.2}) circle (2pt);
    \fill[blue] (1.2, {boundary(1.2)-0.2}) circle (2pt);
    \fill[blue] (2.2, {boundary(2.2)+0.2}) circle (2pt);
    \fill[blue] (3.0, {boundary(3.0)+0.2}) circle (2pt);
    \fill[blue] (2.5,1.2) circle (2pt);
    \fill[blue] (3.1,1) circle (2pt);

    \node[font=\small] at (2,3.3) {Boundary-Aware Subdivision};

    \node[orange] at (1.4,0.5) {\scriptsize Unk};
    \node[green] at (1.35, 2) {\scriptsize Safe};
    \node[orange, scale=0.7] at (3.45,0.5) {Unsafe};
    \node[orange, scale=0.7] at (3.4,2) {Unk};

\end{scope}


\begin{scope}[yshift=-0.8cm]
    \node[scale=0.8] at (1,0) {\textcolor{red}{\textbf{—}} Boundary};
    \node[scale=0.8] at (4.5,0) {\textcolor{blue}{\textbf{- -}} Splits};
    \node[scale=0.8] at (7.5,0) {\textcolor{blue}{\textbf{•}} Samples};
\end{scope}

\end{tikzpicture}
\caption{Uniform vs. boundary-aware subdivision.}
\label{fig:boundary_importance}
\end{figure}
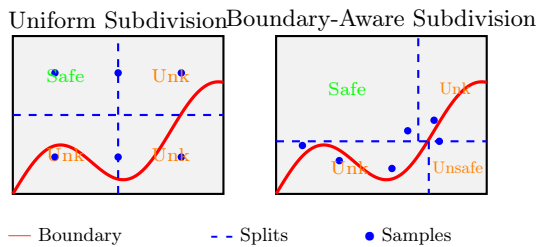

Efficiently finding large safe/unsafe hulls plays a key role in our framework. A naive way could be uniformly subdivide the input state space and verify the safety of each piece. However, it is often very expensive and ignores the decision boundary. Directly considering the boundary in probabilistic hull search can be difficult, since the boundary is in the output space and the subdivision in performed in the input space. However, it is still possible to ``backpropagate'' the information of the boundary to the subdivision process.
Figure~\ref{fig:boundary_importance} gives an intuitive comparison between uniform subdivision and boundary-aware subdivision. The red curve indicates the backpropagated boundary to the input space, it is often not computable and unknown to our process. It can be seen from the figure that, using boundary-aware subdivision can better produce large hulls which do not intersect the boundary and are pure safe or unsafe.




Our boundary-aware subdivision method is outlined in Algorithm \ref{algo:Subdivision}. Given a DNN $f$, a region $R$ in the input space, number of samples $n$ and the safety specification defined by the parameters $\mathbf{C}$ and $\mathbf{a}$. Our whole \emph{boundary-ware subdivision} process performs the following steps: (1) It samples a point set $S_{\text{base}}$ with $n$ points within the given region $R$ and check the safety of their outputs; (2) If both safe and unsafe outputs are found, it conducts boundary-aware sampling by rejecting the samples whose outputs are sufficiently far from the boundary, and we obtain a point set $S$; (3) Otherwise, we set $S = S_{\text{base}}$ since the boundary might not be in $R$; (4) We use the samples in $S$ to build a regression tree $T$ in the region $R$, where each leaf represents a subdivision of $R$. The details of boundary-aware sampling are given as below.

\paragraph{Combined Sampling Strategy Foundation.} We introduce the setup of our base sampling strategy. Our base sampling strategy is the combination of the distributional sampling of the input distribution $\mathcal{N}(\mu, \Sigma)$ and the uniform distribution. Given an integer $n$, we sample $\lfloor n/2 \rfloor$ points with distributional sampling and sample $ n-\lfloor n/2 \rfloor$ points with uniform sampling.
We adopt the combination instead of only one of them because (i) we need to take the distribution of the input variables into account since it helps to find probabilistic hulls in the space of high probability density and make the hulls of high probability, while (ii) additionally considering a uniform distribution helps us to sample the margin of a region, and have balanced safe and unsafe samples.


\paragraph{Boundary Awareness Through Elimination.}
In order to better reflect the safe boundary in $R$, we check our samples to see if both safe and unsafe points are generated. If not, we do not start the rejection sampling step. Otherwise, we aim at keeping the samples near the boundary. Although it is difficult to compute the boundary in the input space, the distance of a sample $\mathbf{x}$ to the boundary can be measured as the distance between $f(\mathbf{x})$ and the boundary. Then we apply an elimination mechanism to remove the samples that are sufficiently far from the boundary. To do so, we associate a probability distribution for eliminating a sample. For each sample $\mathbf{x}$, we compute its output $f(\mathbf{x})$ and the distance to the safety boundary $\mathbf{C}f(\mathbf{x})-\mathbf{a}$. Then, the elimination probability is given by
\[
 \mathbb{P}_e(\mathbf{x}) = 1-\exp(-\text{rank}(|\mathbf{C}·f(\mathbf{x}) - \mathbf{a}|/\sigma))
\]
where $\text{rank}$ returns the corresponding index in the sorted distance from the nearest to the farthest.

\begin{algorithm}[H]
\SetAlgoLined
\KwIn{Region $R$, number of samples $n$, network $f$, $\mathbf{C}$, $\mathbf{a}$}
\KwOut{Boundary-aware regression tree $T$}

\tcp{Combined sampling foundation}

$S_{\text{base}} \leftarrow$CombinedSample($R,n$);

\tcp{Check if rejection sampling is needed}

\If{$S_\text{base}$ contains only safe or unsafe points}{

    \tcp{No boundary in region, return base samples}

    \Return{$S_{\text{base}}$}
}

\tcp{Rejection sampling}
$S \leftarrow \emptyset$\;

$n_{\text{attempts}} \leftarrow 0$\;

\While{$|S| < n$ \textbf{and} $n_{\text{attempts}} < \text{max\_attempts}$}{

    $\mathbf{x} \leftarrow$ CombinedSample($R,n$)

    $\mathbf{x}_{\text{accept}} \leftarrow $EliminateSamples($\mathbf{C}, f,\mathbf{x}$)

    $S \leftarrow S \cup \{\mathbf{x}_{\text{accept}}\}$\;

    $n_{\text{attempts}} \leftarrow n_{\text{attempts}} + 1$\;
}

\tcp{Fill remaining samples if not enough sample points}
\If{$|S| < n$}{
    $S \leftarrow S \cup$ CombinedSample($R$, $n - |S|$)\;
}

$T \leftarrow$  BuildRegressionTree($S$, $R$)

\Return{$T$}
\caption{Boundary-Aware Subdivision}
\label{algo:Subdivision}
\end{algorithm}



\paragraph{Regression Tree Construction.}
We use the sample set after elimination to build a regression tree which is expected to better split the region along the boundary. The tree construction follows the standard algorithm~\cite{Mitchell/ML}.
For each non-leaf node, the impurity of the samples is measured as
\[
 \text{Impurity} = \frac{\text{MSE}(S_{\text{left}}, S_{\text{right}})}{L^\alpha}
\]
where $S_{\text{left}}, S_{\text{right}}$ are the split sets, $L$ is the variation of the samples in the dimension of subdivision, and $\alpha$ is a user-specified hyperparameter controlling the geometric penalty. The MSE is computed as $\text{MSE}(S_{\text{left}}, S_{\text{right}}) = \text{Var}(y_i | x_i \in S_{\text{left}}) \cdot |S_{\text{left}}| + \text{Var}(y_i | x_i \in S_{\text{right}}) \cdot |S_{\text{right}}|$. Each subdivision is performed to minimize the impurity of the two subdivisions.




\begin{algorithm}[H]
\SetAlgoLined
\KwIn{Network $f$, $\mathbf{C}$, $ \mathbf{a}$, distribution $p(\mathbf{x})$, threshold $\varepsilon$}
\KwOut{Safe probability interval $[L_s,U_s]$}

Initialize: $R_{\text{unknown}} \leftarrow$ [entire\_input\_space]\;

$R_{\text{safe}} \leftarrow \emptyset$, $R_{\text{unsafe}} \leftarrow \emptyset$\;

\While{$\sum_{R \in R_{\text{unknown}}} \mathbb{P}(R) \geq \varepsilon$}{
    \tcp{Select region with maximum probability mass}
    $R_{\text{next}} \leftarrow$ SelectLargestProbRegion($R_{\text{unknown}}$);

    \tcp{Targeted sampling within selected region}
    $S \leftarrow$ BoundaryAwareSample($R_{\text{next}}$, small\_sample\_size)\;

    \tcp{Build regression tree and extract leaf regions}
    $T \leftarrow$ BuildRegressionTree($S$, $R_{\text{next}}$)\;

    $\{R_1, ..., R_k\} \leftarrow$ ExtractLeafRegions($T$)\;

    \tcp{Verify each leaf region}
    $\{s_1, ..., s_k\} \leftarrow$ CROWNVerify($\{R_1, ..., R_k\}$, $\mathbf{C}$, $\mathbf{a}$)\;

    \tcp{Update region sets}
    Remove $R_{\text{next}}$ from $R_{\text{unknown}}$\;

    \For{$i \leftarrow 1$ \KwTo $k$}{
        \uIf{$s_i =$ Safe}{
            $R_{\text{safe}} \leftarrow R_{\text{safe}} \cup \{R_i\}$\;
        }
        \uElseIf{$s_i =$ Unsafe}{
            $R_{\text{unsafe}} \leftarrow R_{\text{unsafe}} \cup \{R_i\}$\;
        }
        \Else{
            $R_{\text{unknown}} \leftarrow R_{\text{unknown}} \cup \{R_i\}$\;
        }
    }
}

$L_s \leftarrow \sum_{R \in R_{\text{safe}}} \mathbb{P}(R)$\;

$U_s \leftarrow 1-\sum_{R \in R_{\text{unsafe}}} \mathbb{P}(R)$\;

\Return{$[L_s,U_s]$}
\caption{The Main Algorithm}
\label{algo:iterative_verification}
\end{algorithm}

Algorithm~\ref{algo:iterative_verification} presents the main algorithm that is a detailed treatment of the workflow in Figure~\ref{fig:workflow}. It operates through an iterative refinement loop that progressively subdivides the input space. Rather than attempting to resolve all unknown hulls equally, we strategically select and refine only the candidates with the highest probability mass among the unknown hulls.

\subsection{Analysis of the Framework}

\paragraph{Correctness of the Result.}
We investigate the correctness of the result from Algorithm~\ref{algo:iterative_verification}. First, all of the probabilistic hulls generated intersect at most on the facets. To see that, each hull is obtained as a leaf of a regression tree which subdivide a rougher hull into sub-hulls whose intersections are at most the facets. Therefore, each probabilistic hull in the set $R_{\text{safe}}$ and $R_{\text{unsafe}}$ has a probability which is independent from the probabilities represented by the other hulls. Next, we show that $U_s$ is an over-estimate of the actual safe probability and $L_s$ is an under-estimate of probability. Due to the conservativeness of the tool CROWN, all hulls in $R_{\text{unsafe}}$ are unsafe and all hulls in $\sum_{R \in R_{\text{safe}}}$ are safe. Since the probabilities of the hulls in $R_{\text{unsafe}}$ are independent, we infer that the DNN is at least $1-U_s = \sum_{R \in R_{\text{unsafe}}} \mathbb{P}(R)$ unsafe, and thereby at most $U_s$ safe. Analogously, the DNN is at least $L_s = \sum_{R \in R_{\text{safe}}} \mathbb{P}(R)$ safe since the hulls are safe and representing independent probability. We also have that $L_s \leq U_s$.




\paragraph{Termination of the Main Algorithm.}
We show that Algorithm~\ref{algo:iterative_verification} will eventually terminate with $U_s - L_s < \varepsilon$. Our main algorithm can be viewed as a new branch-and-bound scheme with probability density and boundary-aware subdivision. During the iterations, the sum of the volumes of the hulls in the unknown set $R_{\text{unknown}}$ becomes smaller and smaller due to the repeated refinement. A proof for the termination can be directly obtained by adapting the termination proof of the standard branch-and-bound method~\cite{Jaulin+/2001/applied_interval_analysis}. The volume of the region defined by the union of all undecided hulls, i.e., the hulls in the set of $R_{\text{unknown}}$, converges to zero when the number of iterations goes to infinity. Hence, the total probability of all hulls in $R_{\text{unknown}}$ will eventually be below the given threshold $\varepsilon$ after a finite number of iterations. It is also the probability that lies out of the confirmed safe and unsafe probability estimates, i.e., $U_s - L_s < \varepsilon$.




\begin{theorem}
 Algorithm~\ref{algo:iterative_verification} terminates with an interval that contains the actual safe probability of the DNN.
\end{theorem}



When Algorithm~\ref{algo:iterative_verification} terminates, the union of the hulls in $R_{\text{unknown}}$ forms an overapproximation of the image of the safety boundary via the inverse mapping of the DNN.

\section{Experiments}


 \subsection{Experimental Setup}
 \noindent\textbf{Platform and Tools. }
Our experiment is conducted on a Linux server running Ubuntu 18.04, equipped with an Intel Xeon E5-2678 v3 processor and a GPU of Nvidia RTX 4090. We also use auto\_LiRPA~\cite{DBLP:conf/nips/XuS0WCHKLH20} library to compute the bounds, and our implementation takes as input neural networks in PyTorch format. The safety verification of probabilistic hulls are performed by CROWN~\cite{Zhang2018EfficientNN}.

Our full source code, scripts, and necessary experimental data are publicly available on GitHub at \url{https://github.com/LIJYann/Regression-Tree-Guided-Probabilistic-Hull-Generation}. The repository corresponds to the code version used for this submission, ensuring reproducibility.

\noindent\textbf{Comparison Baselines. }
We compare our approach with two state-of-the-art tools, ProbStar~\cite{QuantitativeVerificationProbStars} and the basic branch-and-bound (BaB) method~\cite{PVBranchBound} (binary partition on the longest dimension). The comparison is done on three groups of DNN benchmarks: ACAS Xu,  the rocket lander controller benchmark for SpaceEx Falcon9, and the $\tanh$-based DNNs distilled from those in ACAS Xu (see~\cite{DBLP:journals/ijcv/GouYMT21}). We use a less restrictive condition, $\max_{R\in R_{\text{unknown}}}\mathbb{P}(R)\leq\varepsilon$, for the termination condition in order to reach a reasonable tradeoff between efficiency and accuracy. This condition is used in both our method and the BaB in order to give a fair comparison. The parallel computation in BaB and our approach are conducted on GPU with auto\_LiRPA. However, due to the implementation of ProbStar, it can only be parallelized on CPU, and we use the default parallel setting for ProbStar. The experimental results from the tools are compared based on the runtime and the tightness of the upper and lower bounds for the actual safe probability.

\noindent\textbf{Hyperparameter Settings. }
To systematically evaluate our regression-tree-guided probabilistic verification method across different parameter configurations, we conducted a comprehensive grid search on each benchmark. We test all combinations of key parameters through their Cartesian product, enabling us to identify optimal configurations. We examined three critical parameters: sampling strategy, regression tree depth, and impurity hyperparameter settings.
Overall, the configurations can be described as $((w_\text{uniform},w_\text{distribution}),\text{depth},(\alpha,\beta))$:
\begin{itemize}
    \item Sampling strategy: Weight configurations balancing uniform and distributional sampling $(w_\text{uniform},w_\text{distribution})$;
    \item Regression tree depth: Depth limits on regression tree;
    \item Impurity scheduler: Combined the proposed $\alpha$-weighted measure with longest-dimension splitting. Initial splits follow the longest dimension until the hull probability reaches threshold $\beta$, then transition to the soft-coded impurity measure $(\alpha,\beta)$.
\end{itemize}
\begin{figure}[htbp]
    \centering
    \includegraphics[width=0.6\linewidth]{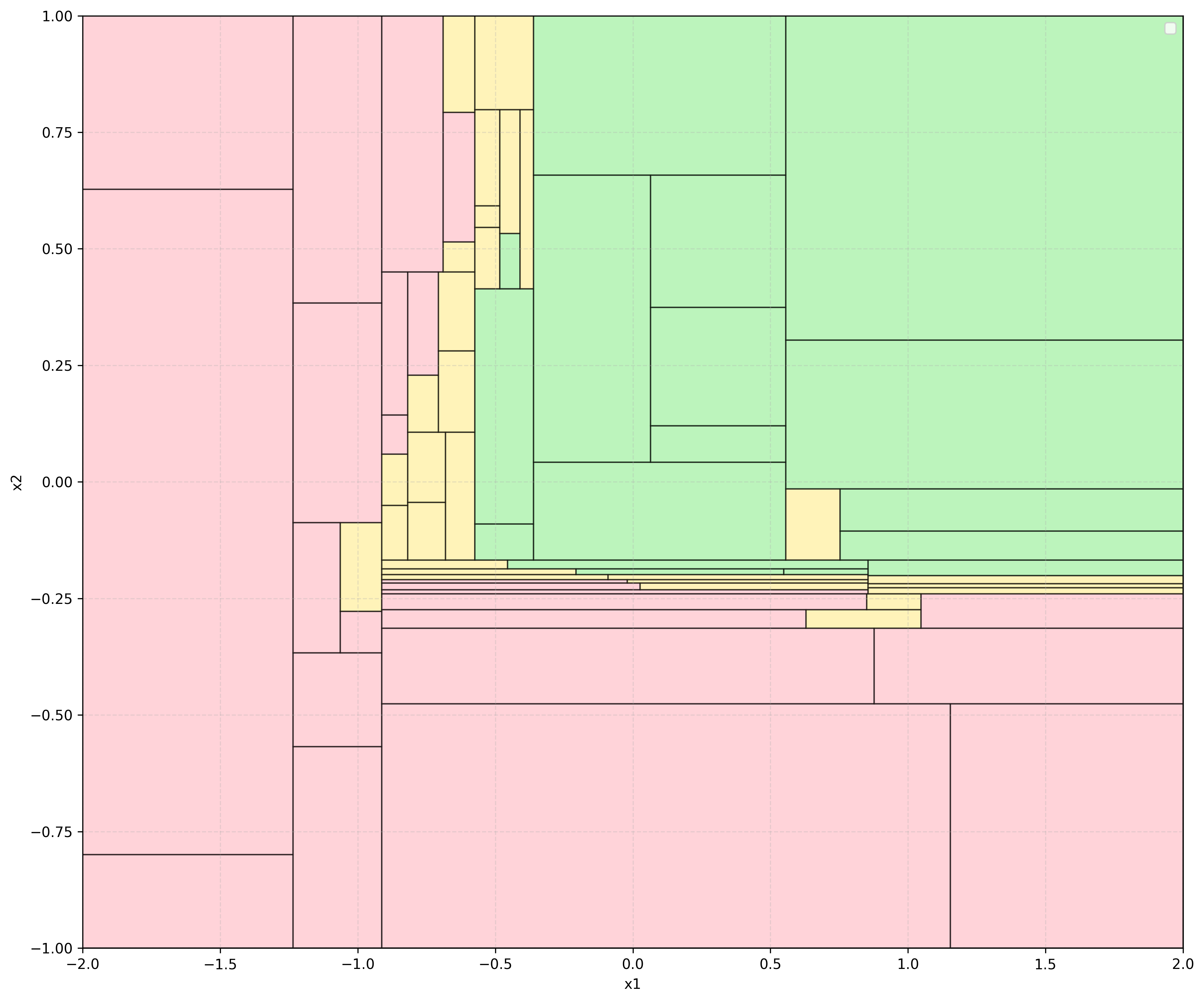}
    \caption{Boundary-aware subdivision for the toy example.}
    \label{fig:toy}
\end{figure}
\noindent\textbf{Illustrative Toy Examples. }
Before we show the comparisons, Figure~\ref{fig:toy} illustrates the probabilistic hulls computed for the toy example described in~\cite{QuantitativeVerificationProbStars}. The red boxes are the probabilistic hulls that are identified as unsafe. The green boxes are the safe hulls. While the yellow ones are the boxes whose corresponding outputs intersect the safety boundary. The union of the yellow boxes forms an overapproximation of the image of the safety boundary under the inverse mapping of the neural network. It can be seen that the state far from the boundary is often included by large probabilistic hulls, i.e., some unnecessary subdivisions can be avoided using our boundary-aware method. The rest of the section addresses the experimental evaluation goals as mentioned previously.




\subsection{ACAS Xu Benchmark}
\begin{table*}[htbp]
\centering
\caption{Results on ACAS Xu Benchmark for Property P2. Legend - \textbf{Net}: DNN benchmarks,
\textbf{Approach}: our approach ("Ours"), ProbStar and BaB,
$L_s$: the lower bound of the safety probability interval, $U_s$: the upper bound of the safety probability interval,
$U_s-L_s$: upper bound for the unknown probability,
Time (non parallel): the runtime without parallel computing in seconds, Time (parallel): the runtime with parallel computing in seconds, T.O.: Time out, $>$2 hours.}
\begin{tabular}{ccccccc}
\toprule
\textbf{Net} & \textbf{Approach} & \textbf{$L_s$} & \textbf{$U_s$} & $U_s-L_s$ & \textbf{Time (non parallel)} & \textbf{Time (parallel)} \\
\midrule
1-6 & Ours& \textbf{0.984714} & 1.000000 & \textbf{0.015286} & \textbf{124.21} & \textbf{46.58}\\
1-6 & ProbStar & 0.933739 & \textbf{0.997192} & 0.063453 & 2840.35& 315.75 \\
1-6 & BaB & -& -& -& T.O.& T.O.\\
\hline
2-2 & Ours& \textbf{0.924122} & 0.991622 & \textbf{0.067500} & \textbf{1248.70} & \textbf{240.79}\\
2-2 & ProbStar & 0.892540 & \textbf{0.980435} & 0.087895 & 3869.90& 430.20 \\
2-2 & BaB & -& -& -& T.O. & T.O.\\
\hline
2-9 & Ours& \textbf{0.954046} & 0.999941 & \textbf{0.045895} & \textbf{819.22} & \textbf{249.24}\\
2-9 & ProbStar & 0.879616 & \textbf{0.999745} & 0.120129 & 6715.75& 746.56 \\
2-9 & BaB&-& -& -&T.O.& T.O.\\
\hline
3-1 & Ours& \textbf{0.923320} & 0.972056 & \textbf{0.048735} & \textbf{949.94} & \textbf{189.76}\\
3-1 & ProbStar & 0.913943 & \textbf{0.969500} & 0.055557 & 2462.36& 273.73 \\
3-1 &BaB & -& -& -& T.O. & T.O.\\
\hline
3-6 & Ours& \textbf{0.914964} & 0.982248 & \textbf{0.067284} & \textbf{1808.11} & \textbf{359.85}\\
3-6 & ProbStar & 0.879708 & \textbf{0.979217} & 0.099509 & 5191.90& 577.16 \\
3-6 &BaB & -& -&-& T.O. &T.O. \\
\hline
3-7 & Ours& \textbf{0.923570} & 0.999888 & \textbf{0.076318} & \textbf{1817.89} & \textbf{386.33}\\
3-7 & ProbStar & 0.911578 & \textbf{0.997681} & 0.086103 & 4187.35& 465.49 \\
3-7 &BaB & -& -& -&T.O. & T.O.\\
\hline
4-1 & Ours& \textbf{0.961700} & 0.999953 & \textbf{0.038253} & \textbf{884.45} & \textbf{155.31}\\
4-1 & ProbStar & 0.914955 & \textbf{0.998963} & 0.084008 & 3120.21& 346.86 \\
4-1 &BaB & -& -& -& T.O. & T.O.\\
\hline
4-7 & Ours& \textbf{0.913432} & 0.979347 & \textbf{0.065914} & \textbf{1604.20} & \textbf{381.73}\\
4-7 & ProbStar & 0.878505 & \textbf{0.979215} & 0.100710 &3917.94& 435.54 \\
4-7 &BaB & -& -& -& T.O. & T.O.\\
\hline
5-3 & Ours& \textbf{0.973373} & \textbf{1.000000} & \textbf{0.026627} & \textbf{336.51} & \textbf{81.21}\\
5-3 & ProbStar & 0.953952 & \textbf{1.000000} & 0.046048 & 1306.79 & 145.27 \\
5-3 & BaB& -& -& -& T.O. & T.O.\\
\bottomrule
\end{tabular}
\label{tab:ACAS}
\end{table*}

The ACAS Xu collision avoidance system benchmarks~\cite{manfredi2016introduction} consist of 45 networks (5×9 configurations) with 5 inputs, 6 layers, and 50 neurons per layer. 
We test Property P2 in the benchmarks which requires the network not to output ``COC'' as the optimal action. This means the network output is \textbf{unsafe} when the first output is the minimal.
In this comparison, all methods filter regions (branches in ProbStar's case) smaller than $10^{-5}$. We use 1000 initial sampling points and 100 iterative sampling points in our approach.

As for the hyperparameters, we performed the grid search on ACAS network 1-6 for specification 2 and determine the optimal configuration to be pure distributional sampling (0.0,1.0), depth 5, $\beta=0.75$, $\alpha=0.05$. We therefore adopt these hyperparameter settings for the ACAS benchmark. The full illustration of the grid search can be viewed in Section~\ref{hyperparameter}.

Table~\ref{tab:ACAS} shows the results for all tools, where the best bounds and runtimes are marked in bold (and for all subsequent tables). From the table, we observe that BaB times out on all the benchmarks, and our method mostly generates better bounds than  ProbStar. Specifically, our bounds uniformly outperforms ProbStar in the bounds for unknown probabilities $U_s-L_s$, and our runtime is comparable with ProbStar. Besides, the parallel paradigm originated from CROWN yields substantial performance gains, with an average speedup of 4.8× compared to our non-parallel version.

\subsection{Rocket Lander Controller Benchmark}
To further demonstrate the ability of our method to handle high-dimensional neural networks, we conduct evaluation on the Rocket Lander~\cite{DBLP:conf/formats/YangYTHJP22}, which simulates the vertical landing control of a SpaceX Falcon 9 first-stage rocket. The objective is to safely land the rocket on a sea-based platform from a given height, which requires precise control of both the rocket's velocity and lateral angle.

The control system outputs three components: a main engine thruster ($F_E \in [0,1]$) with an adjustable angle ($\phi$), and two side nitrogen thrusters ($F_S \in [-1,1]$, where -1 and 1 indicate full throttle of right and left thruster respectively). The system state is represented by a 9-dimensional vector: $(x, y, v_x, v_y, \theta, \omega, F'_E, F'_S, \phi')$, where $(x,y)$ is the position relative to the landing center; $(v_x,v_y)$ are the horizontal and vertical velocities; $\theta$ is the lateral angle; $\omega$ is the angular velocity; and $(F'_E, F'_S, \phi')$ are the previous control actions.

We directly use the control policy network provided in ProbStar. The network has 9 inputs, 3 outputs, and 5 hidden layers containing 20 ReLU neurons each. We focus on two critical safety properties that ensure proper tilt control:

\begin{itemize}
    \item $P_1$ (Right Tilt Prevention): For states where $\theta \in [-20\degree, -6\degree]$, $\omega < 0$, $\phi' \leq 0$, and $F'_S \leq 0$, the network must output either $\phi < 0$ or $F_S < 0$.
    \item $P_2$ (Left Tilt Prevention): For states where $\theta \in [6\degree, 20\degree]$, $\omega \geq 0$, $\phi' \geq 0$, and $F'_S \geq 0$, the network must output either $\phi > 0$ or $F_S > 0$.
\end{itemize}

As for the hyperparameters, we performed the grid search on Net 1 for specification 1 and determine the optimal configuration to be mixed sampling $(w_\text{uniform}=0.25,w_\text{distribution}=0.75)$, depth 5, $\beta=0.0$, $\alpha=0.05$. We therefore adopt these hyperparameter settings for this benchmark. The scatter point figure of the grid search can be viewed in Section~\ref{hyperparameter}.

We conduct experiments on the Rocket Lander controller to compare our method with ProbStar. In this comparison, all methods filter regions (branches in ProbStar's case) smaller than $10^{-3}$.  We do not run BaB since it cannot handle high-dimensional input due to inherent combinatorial explosion. For our method, we use 9000 initial sampling points and 900 iterative sampling points to adapt to the high input dimension of this benchmark. Similar to the setting in Table~\ref{tab:ACAS}, we compare both the bounds and the runtime (including parallel and non-parallel). The results are shown in Table~\ref{tab:results_table} (non-parallel) and Table~\ref{tab:parallep results on rocketnet} (parallel). From these tables, we observe that our method consistently generates significantly tighter bounds for the unknown probability ($U_s - L_s$) across all configurations. Furthermore, our method demonstrates superior time efficiency, achieving substantial runtime reductions in both non-parallel and parallel settings.

\begin{table}[htbp]
    \centering
     \caption{Experiments on Rocket Lander Controller (non-parallel). \textbf{Net} \& \textbf{Prop} means different configurations in the original benchmark. Others are the same as Table~\ref{tab:ACAS}.}
    \begin{tabular}{cccc}\toprule
         \textbf{Net} \& \textbf{Prop}& \textbf{Approach} &  $U_s-L_s$& \textbf{Time (s)} \\\midrule
         0 \& 1 & ProbStar &  0.715046& 418.183 \\
         0 \& 1 & Ours & \textbf{0.650548}& \textbf{171.933}\\ \midrule
         0 \& 2 & ProbStar & 0.831752& 597.775\\
         0 \& 2 & Ours & \textbf{0.262993}& \textbf{85.482}\\ \midrule
         1 \& 1& ProbStar & 0.554032& 438.278 \\
         1 \& 1& Ours & \textbf{0.229874}& \textbf{70.723}\\ \midrule
         1 \& 2& ProbStar & 0.356913& 319.571 \\
         1 \& 2& Ours & \textbf{0.079113}& \textbf{23.287}\\
         \bottomrule
    \end{tabular}

    \label{tab:results_table}
\end{table}

\begin{table}[htbp]
    \centering
     \caption{Experiments on Rocket Lander Controller (parallel). The legend is the same as Table~\ref{tab:results_table}.}
    \begin{tabular}{cccc}\toprule
         \textbf{Net} \& \textbf{Prop}& \textbf{Approach} &  $U_s-L_s$ & \textbf{Time (s)} \\\midrule
         0 \& 1 & ProbStar & 0.715046 & 141.047 \\
         0 \& 1 & Ours & \textbf{0.651642} & \textbf{33.767}\\ \midrule
         0 \& 2 & ProbStar & 0.831752 & 225.331\\
         0 \& 2 & Ours &\textbf{0.265237} & \textbf{23.573}\\ \midrule
         1 \& 1& ProbStar & 0.554032 & 225.331\\
         1 \& 1& Ours & \textbf{0.230962} & \textbf{18.900}\\ \midrule
         1 \& 2& ProbStar & 0.356913 & 117.494\\
         1 \& 2& Ours & \textbf{0.079063} & \textbf{12.947}\\
         \bottomrule
    \end{tabular}

    \label{tab:parallep results on rocketnet}
\end{table}

\subsection{Efficiency of Finding Probabilistic Hulls}

We compare the efficiency of our method and the BaB on finding safe and unsafe probabilistic hulls, since both of the methods use probabilistic hulls to estimate the safe probability. We use the distilled DNNs~\cite{DBLP:journals/ijcv/GouYMT21} of the ones in the ACAS Xu set. All of the new DNNs have $\tanh$ activation function. Moreover, we consider the parallel setting for both our method and BaB. The hyperparameters are the same as in the original ACAS Xu experiments.

The comparison is made in the following way. We set up an end time of 1800 seconds to let both of the methods find probabilistic hulls. Each of the methods may terminate either at the end time or when the early termination condition holds (i.e., the highest probability of the undetermined sets is less than the given threshold $\varepsilon$, similar to $\max_{R\in R_{\text{unknown}}}\mathbb{P}(R)\leq\varepsilon$). Then we compare the runtime and the estimated upper and lower bounds for the safe probability. Clearly, the tighter the bounds, the better the result.

We use $\varepsilon = 10^{-5}$, and the experimental results are given in Table~\ref{tab:acas_results}. It can be seen that our method always stops before the end time and the runtime are much lower than the basic BaB method. On the other hand, the bounds obtained from our method are also much tighter than those from the basic BaB ones. Hence, our method is much more efficient in finding safe and unsafe probabilistic hulls.
\begin{table}[htbp]
\centering
\small
\caption{Efficiency of Finding Probabilistic Hulls (parallel). Legend - Time: runtime in seconds, End: end time = 1800 seconds, Others are the same as those in Table~\ref{tab:ACAS}.}
\label{tab:acas_results}
\begin{tabular}{c|c|ccc|c}\toprule
\textbf{Net} & \textbf{Approach} & \textbf{$L_s$} & \textbf{$U_s$} & \textbf{$U_s-L_s$} & \textbf{Time (s)} \\\midrule
1-6 & Ours & \textbf{0.9815} & \textbf{0.9995} & \textbf{0.0180} & \textbf{264.00} \\
1-6 & BaB & 0.9397 & 1.0000& 0.0603 & End \\
\hline
2-2 & Ours &\textbf{ 0.9819} & \textbf{0.9997}& \textbf{0.0178} & \textbf{142.30} \\
2-2 & BaB & 0.9616 & 1.0000 & 0.0384 & End \\
\hline
2-9 & Ours &\textbf{0.9502} & \textbf{0.9994}& \textbf{0.0492} & \textbf{686.66} \\
2-9 & BaB & 0.8896 & 1.0000& 0.1104 & End \\
\hline
3-1 & Ours & \textbf{0.9865} & 1.0000& \textbf{0.0135} &\textbf{ 68.06} \\
3-1 & BaB & 0.9845 & 1.0000& 0.0155 & 1413.22 \\
\hline
3-6 & Ours & \textbf{0.9293} & \textbf{0.9976}& \textbf{0.0682} & \textbf{1188.63} \\
3-6 & BaB & 0.7738 & 1.0000& 0.2262 & End \\
\hline
3-7 & Ours & \textbf{0.8673} &  \textbf{0.9448} & \textbf{0.0775} &\textbf{1338.55} \\
3-7 & BaB & 0.7682 &  1.0000& 0.2318 & End \\
\hline
4-1 & Ours & \textbf{0.9864} &  1.0000& \textbf{0.0136} & \textbf{59.52} \\
4-1 & BaB & 0.9837 &  1.0000& 0.0163 & 1019.67 \\
\hline
4-7 & Ours & \textbf{0.9778} &  1.0000& \textbf{0.0222} & \textbf{397.20} \\
4-7 & BaB & 0.8993 &  1.0000& 0.1007 & End \\
\hline
5-3 & Ours &\textbf{0.9823} &  1.0000& \textbf{0.0177} & \textbf{303.09} \\
5-3 & BaB & 0.8886 &  1.0000& 0.1114 & End \\ \bottomrule
\end{tabular}
\end{table}

\subsection{Discussion}

Basically, our method can handle most of the feedforward DNNs with various activation functions, since we use the given DNN as a blackbox to produce the samples. This is a clear advantage over the state-of-the-art tools such as ProbStar which requires neural networks only to have ReLU activation functions. On the other hand, our boundary-aware subdivision demonstrate a great performance improvement in not only time efficiency but also accuracy comparing to the basic branch-and-bound method.

In the experiments, we also find two weaknesses of our method. First, the total runtime highly depends on the time spent by CROWN which is used to verify the safety of a hull in the input space. It can be improved by designing a more efficient output range analysis technique which is particularly for the hulls in our problem. Second, our method still suffers from the curse of dimensionality in the worst case, although the boundary-aware subdivision helps a lot on avoiding subdividing the non-critical regions. This aspect can be improved by considering symbolic representations of the hulls, such as Taylor models~\cite{DBLP:conf/atva/HuangFCLZ22}.

Besides, we observed that some hulls only have a very small intersection with the boundary but requires many subdivisions to find purely safe or unsafe subsets. For such a hull, we consider to develop shrinking methods that can fast find subsets by contracting the original hull.

\section{Conclusion}

We present a novel approach of probabilistic verification of DNNs and the main idea is to efficiently find safe or unsafe probabilistic hulls using a boundary-aware subdivision strategy. Our method does not have a limitation to particular activation functions and shows a much better performance than the similar existing techniques.

In the future, we seek to improve the method by (i) developing more efficient output range computation techniques particularly for the probabilistic hulls, to make the verification step less time-consuming; (ii) designing new symbolic representations for the hulls such that they are not limited to boxes or parallelotopes.
\bibliographystyle{splncs04}

\bibliography{mybibliography}

\appendix
\section{Hyperparameter and Grid Search}\label{hyperparameter}
We performed an extensive grid search to determine the best configuration for the core components of our approach: the sampling strategy, the impurity scheduler, and the regression tree depth.
\subsection{ACAS Xu Benchmark}
For the ACAS Xu benchmark, the search space was defined as follows.

\begin{itemize}
    \item \textbf{Sampling Weights ($w_{\text{uniform}}, w_{\text{distribution}}$):} Five configurations were tested. $(0.0, 1.0)$, $(0.25, 0.75)$, $(0.5, 0.5)$, $(0.75, 0.25)$, and $(1.0, 0.0)$.
    \item \textbf{Regression Tree Depth:} Two values were considered. $5$ and $10$.
    \item \textbf{Impurity Splitting ($\beta$):} Five values were tested. $0, 0.25, 0.5, 0.75$, and $1.0$.
    \item \textbf{Impurity Coefficient ($\alpha$):} Six values were evaluated. $0.05, 0.1, 0.3, 0.5, 1.0$, and $2.0$.
\end{itemize}
This parameter space resulted in $\mathbf{25}$ impurity configurations and a total of $\mathbf{250}$ parameter combinations. The evaluation metrics for each combination were the remaining unknown region probability ($U_s-L_s$) and the corresponding runtime.

\begin{figure}[htbp]
    \centering
    \includegraphics[width=0.8\linewidth]{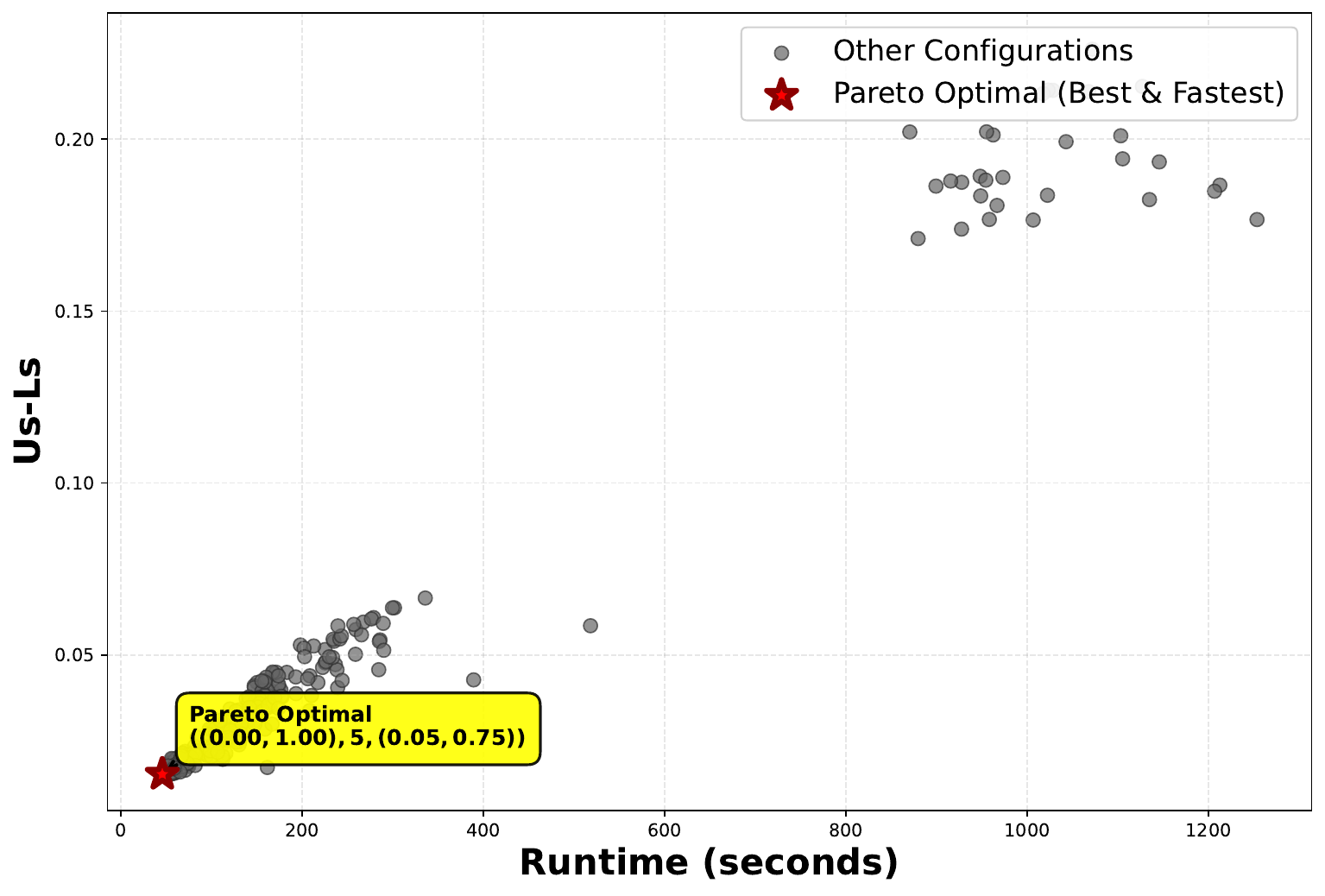}
    \caption{Scatter plot showing the relationship between Us-Ls and runtime for 250 different parameter configurations tested in the grid search. }
    \label{fig:grid_acas}
\end{figure}

The grid search was performed on \textbf{ACAS network 1-6 for specification 2}. Figure~\ref{fig:grid_acas} presents the results, mapping the trade-off between $U_s-L_s$ (vertical axis) and runtime (horizontal axis). Each point represents a distinct configuration. The red star highlights the \textbf{Pareto optimal configuration}, which outperforms all other combinations in both metrics. Gray points indicate configurations that are dominated by the Pareto solution in at least one metric.

The search identified the optimal configuration for the ACAS benchmark as pure distributional sampling $(0.0, 1.0)$, a regression tree depth of 5, $\beta=0.75$, and $\alpha=0.05$. These settings were subsequently adopted for the main experiments on the ACAS Xu benchmark.

\subsection{Rocket Lander Benchmark}
\begin{figure}[htbp]
    \centering
    \includegraphics[width=0.8\linewidth]{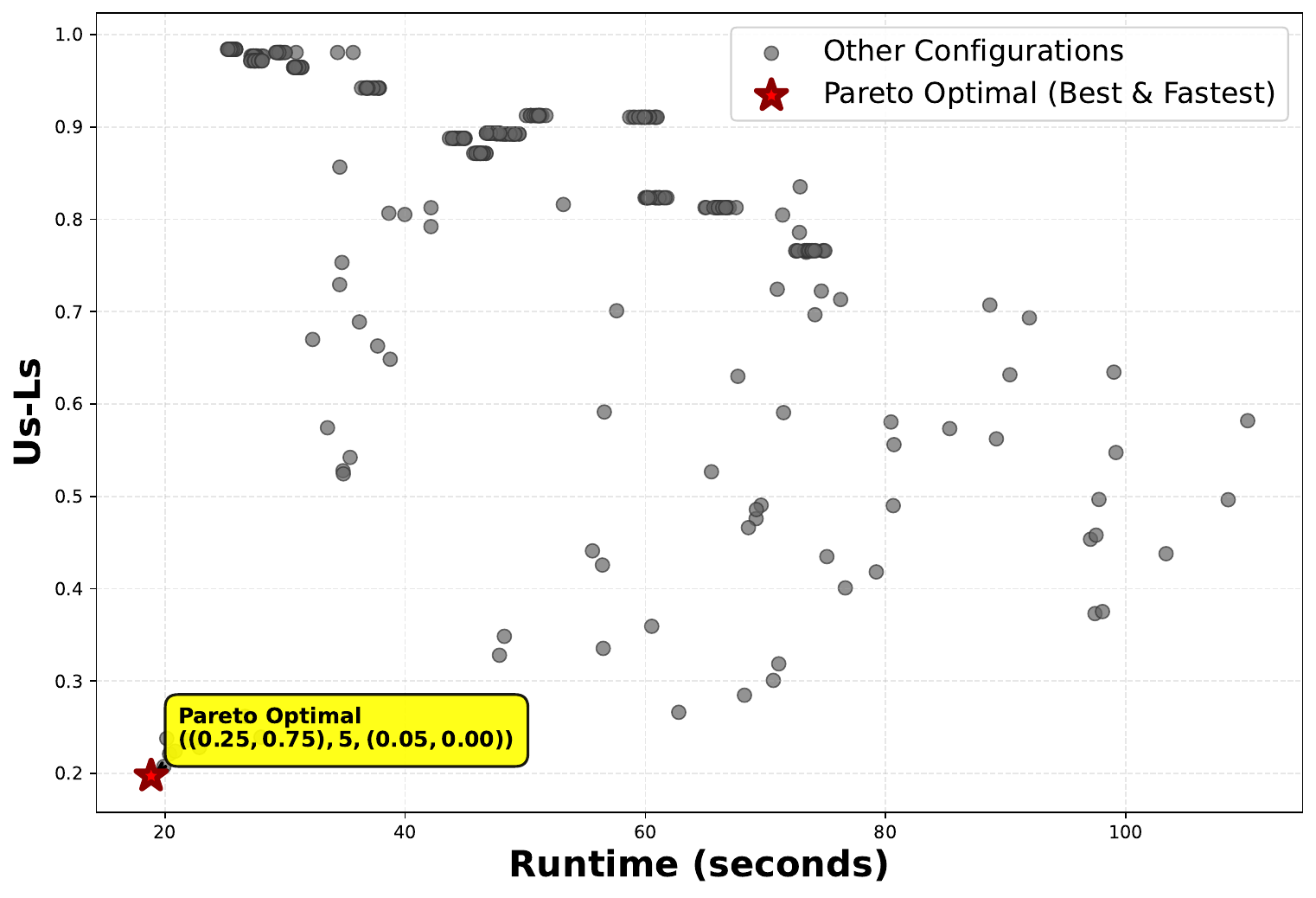}
    \caption{Scatter plot showing the relationship between Us-Ls and runtime for 300 different parameter configurations tested in the grid search. }
    \label{fig:grid_rknet}
\end{figure}
The same systematic grid search methodology was applied to the Rocket Lander controller. The search space for this benchmark was defined as follows.

\begin{itemize}
    \item \textbf{Sampling Weights ($w_{\text{uniform}}, w_{\text{distribution}}$). } The same five configurations were tested. $(0.0, 1.0)$, $(0.25, 0.75)$, $(0.5, 0.5)$, $(0.75, 0.25)$, and $(1.0, 0.0)$.
    \item \textbf{Regression Tree Depth. } Three values were considered. $5$, $10$, and $20$.
    \item \textbf{Impurity Splitting ($\beta$). } Five values were tested. $0, 0.25, 0.5, 0.75$, and $1.0$.
    \item \textbf{Impurity Coefficient ($\alpha$). } Five values were evaluated. $0.05, 0.1, 0.3, 0.5$, and $1.0$.
\end{itemize}
This yielded a total of $\mathbf{300}$ parameter combinations. The grid search was performed on \textbf{Rocket Net 1 for specification 1}.

Figure~\ref{fig:grid_rknet} presents the corresponding Pareto front analysis. Consistent with the ACAS results, the $U_s-L_s$ is plotted against the runtime, and the red star marks the Pareto optimal configuration.

The optimal configuration determined for the Rocket Lander benchmark was a *mixed sampling strategy $(0.25, 0.75)$, a regression tree depth of 5, $\beta=0.0$, and $\alpha=0.05$**. These settings were adopted for the main experiments on the Rocket Lander benchmark.

\section{Ablation Studies}
We conduct a comprehensive ablation study to analyze the independent contribution and combinatorial effects of the key components in our proposed approach. The components investigated include the weighted mixed sampling and the impurity scheduler defined by $\alpha$ and $\beta$ in our hyperparameter settings. The ablation studies are conducted on both the high-dimensional Rocket Lander and the moderate-dimensional ACAS Xu benchmarks. The following sections detail the findings, quantifying the impact of each mechanism on verification tightness ($U_s - L_s$) and time efficiency across these diverse benchmarks.

\subsection{Contribution of Sampling and Impurity Scheduling}

\begin{table*}[t]
    \centering
    \caption{Ablation Study on Rocket Lander and ACAS Xu (Average Results). The table compares the average unknown probability ($U_s-L_s$) and average runtime for different component configurations on two benchmarks. The weights are $w_u$ (uniform) and $w_d$ (distributional).}
    \label{tab:ablation_dual_benchmark}
    \resizebox{\textwidth}{!}{
    \begin{tabular}{l|c|cc|cc}
    \toprule
    \multirow{2}{*}{\textbf{Configuration}} & \multirow{2}{*}{\textbf{($w_u, w_d$)}} & \multicolumn{2}{c|}{\textbf{Rocket Lander}} & \multicolumn{2}{c}{\textbf{ACAS Xu}} \\
    \cmidrule(lr){3-4} \cmidrule(lr){5-6}
    & & \textbf{Avg. $U_s - L_s$} & \textbf{Avg. Time (s)} & \textbf{Avg. $U_s - L_s$} & \textbf{Avg. Time (s)} \\
    \midrule
    \multicolumn{6}{l}{\textbf{Baseline Configurations (Impurity Scheduler Disabled)}} \\
    \midrule
    \multirow{5}{*}{Baseline} & (1.00, 0.00) & 0.9194 & 30.26 & 0.0650 & 279.46 \\
    & (0.00, 1.00) & 0.9242 & 24.86 & 0.0681 & 279.05 \\
    & (0.75, 0.25) & 0.9134 & 30.74 & 0.0652 & 277.95 \\
    & (0.50, 0.50) & 0.9124 & 29.22 & 0.0664 & 279.09 \\
    & (0.25, 0.75) & 0.9124 & 27.80 & 0.0662 & 278.14 \\
    \midrule
    \multicolumn{6}{l}{\textbf{Configurations with Impurity Scheduler Enabled}} \\
    \midrule
    \multirow{5}{*}{+ Impurity Scheduler} & (1.00, 0.00)$^*$ & 0.2466 & 23.17 & 0.0341 & 211.61 \\
    & (0.00, 1.00) & 0.3294 & 25.23 & 0.0368 & 221.05 \\
    & (0.75, 0.25) & \textbf{0.2371} & \textbf{21.25} & \textbf{0.0340} & \textbf{210.33} \\
    & (0.50, 0.50) & 0.2705 & 22.02 & 0.0344 & 214.28 \\
    & (0.25, 0.75) & 0.2576 & 21.71 & 0.0350 & 218.96 \\
    \bottomrule
    \end{tabular}
    }
    \footnotesize
    $^*$For ACAS Xu, the impurity scheduler uses $\beta=0.75$ and $\alpha=0.05$ as the representative Impurity-only configuration for comparison. For Rocket Lander, the impurity scheduler uses $\beta=0.0$ and $\alpha=0.05$ as the representative Impurity-only configuration for comparison.
\end{table*}

The analysis of results presented in Table~\ref{tab:ablation_dual_benchmark} focuses on the core question of whether sampling alone is effective and how the Impurity Scheduler contributes to the final performance. The data reveals that the \textbf{Impurity Scheduler} is the critical component for achieving high verification accuracy ($U_s - L_s$).

When the Impurity Scheduler is disabled, varying the sampling ratio only provides marginal improvements in $U_s - L_s$. For the Rocket Lander, the best sampling ratio ($w_u=0.25$ or $w_u=0.50$) achieves $0.9124$, barely better than the worst sampling ratio ($0.9194$). For ACAS Xu, the sampling variance shows negligible impact on accuracy.

However, enabling the Impurity Scheduler changes the performance profile. Firstly, for the pure uniform sampling case ($w_u=1.00$), $U_s - L_s$ is reduced from $0.9194$ to $0.2466$ for Rocket Lander and from $0.0650$ to $0.0341$ for ACAS Xu. This confirms that the scheduler can significantly tighten the verification bounds.

Furthermore, when the Impurity Scheduler is active, optimizing the mixed sampling ratio becomes relevant, providing a measurable gain. For both benchmarks, the $\mathbf{(0.75, 0.25)}$ mixed ratio achieves the tightest overall bound ($0.2371$ for Rocket Lander and $0.0340$ for ACAS Xu). This indicates that while sampling is not the main driver of accuracy, an optimal mix of uniform and distributional sampling is necessary for the final refinement.

Regarding runtime efficiency, the runtime for nearly all configurations, remains within a similar order of magnitude on both benchmarks. For instance, the Rocket Lander configurations generally fall between $21\text{s}$ and $31\text{s}$, and the ACAS Xu configurations are mostly between $210\text{s}$ and $280\text{s}$. This suggests that the major performance contribution of the Impurity Scheduler lies in the massive improvement of \textbf{accuracy} ($U_s - L_s$) convergence, while overall execution time remains relatively constant across effective configurations.

\subsection{Ablation on Impurity Scheduler Threshold ($\beta$)}
\begin{table*}[t]
    \centering
    \caption{Ablation Study on Impurity Scheduler Threshold ($\beta$). The table compares the performance ($U_s-L_s$ and Runtime) of the Impurity Scheduler by varying the threshold $\beta$, which controls the transition from longest-dimension splitting ($\beta=1.0$) to soft-coded impurity splitting ($\beta < 1.0$). Only pure uniform sampling ($w_u=1.00$) is used.}
    \label{tab:ablation_beta_comparison}
    \resizebox{\textwidth}{!}{
    \begin{tabular}{c|c|cc|cc}
    \toprule
    \multirow{2}{*}{\textbf{Threshold $\beta$}} & \multirow{2}{*}{\textbf{Splitting Mode}} & \multicolumn{2}{c|}{\textbf{Rocket Lander}} & \multicolumn{2}{c}{\textbf{ACAS Xu}} \\
    \cmidrule(lr){3-4} \cmidrule(lr){5-6}
    & & \textbf{Avg. $U_s - L_s$} & \textbf{Avg. Time (s)} & \textbf{Avg. $U_s - L_s$} & \textbf{Avg. Time (s)} \\
    \midrule
    \textbf{Baseline ($\beta$ Implied)} & Longest-Dimension & 0.9194 & 30.26 & 0.0650 & 279.46 \\
    \midrule
    $0.75$ & Late Activation & 0.9194 & 30.24 & \textbf{0.0341} & 211.61 \\
    $0.50$ & Mid Activation & 0.9194 & 30.53 & 0.0342 & \textbf{195.23} \\
    $0.25$ & Early Activation & 0.9194 & 30.29 & 0.0378 & 210.71 \\
    $0.00$ & Immediate Soft Impurity & \textbf{0.2466} & \textbf{23.17} & 0.0610 & 349.11 \\
    \bottomrule
    \end{tabular}
    }
\end{table*}
Given the critical role of the Impurity Scheduler, we conduct a focused ablation study by varying the threshold $\beta$, which controls the strategy for transitioning from standard longest-dimension splitting to the proposed soft-coded impurity measure ($\alpha$-weighted splitting). The results, using pure uniform sampling ($w_u=1.00$), are summarized in Table~\ref{tab:ablation_beta_comparison}. The optimal transition strategy is found to be benchmark-dependent.

For Rocket Lander benchmark, the critical finding is that the \textbf{Immediate Soft Impurity ($\beta=0.00$)} mode is optimal. This configuration achieves the tightest bound of $\mathbf{0.2466}$ and the best runtime of $\mathbf{23.17\text{s}}$ within this set. This suggests that the longest-dimension splitting provides minimal structural benefit, and immediate application of the boundary-aware measure is key.

As for ACAS Xu, a \textbf{hybrid splitting strategy} is essential. The Immediate Soft Impurity ($\beta=0.00$) yields a poor $U_s - L_s$ of $0.0610$ and the longest runtime of $349.11\text{s}$. In contrast, the best accuracy ($\mathbf{0.0341}$) is achieved when $\beta=0.75$ (Late Activation), allowing the longest-dimension splitting to run for the initial portion of the verification. The best runtime ($\mathbf{195.23\text{s}}$) is achieved at $\beta=0.50$ (Mid Activation). This confirms that for ACAS Xu, initial global refinement via standard splitting significantly accelerates the process before the fine-grained, boundary-aware splits are activated.

The distinct optimal $\beta$ values can be explained by the fundamental differences in input dimensionality and the subsequent geometric challenges presented to the verification method (which relies on convex hull approximations):
\begin{enumerate}
    \item \textbf{Geometric Control:} In methods utilizing convex hull approximations, the Longest-Dimension splitting is employed primarily to maintain a \textbf{low inter-dimension ratio} for the divided regions. Preventing regions from becoming highly elongated is vital, as this mitigates the over-approximation error that occurs when hull methods are applied to non-square regions.
    \item \textbf{High Dimensionality (Rocket Lander):} The 9-dimensional space of Rocket Lander limits the effectiveness of pure Longest-Dimension splitting. The soft impurity measure from the beginning is necessary because its geometric penalty term ($L^\alpha$) simultaneously guides the split toward reducing boundary uncertainty (via MSE) and acts to regularize the region's aspect ratio by penalizing splits across large dimensions ($L$). This combined efficiency makes immediate activation optimal, as the structural splitting stage provides little benefit.
    \item \textbf{Initial Refinement Need (ACAS Xu):} In the lower 5-dimensional ACAS Xu space, the Longest-Dimension splitting remains an efficient method for achieving an initial, coarse, global subdivision. This initial phase (enabled by $\beta=0.75$ or $\beta=0.50$) establishes regions with more uniform aspect ratios. This preprocessing ensures that when the fine-grained soft impurity measure is activated later, it operates on geometrically well-behaved regions, thereby improving both accuracy and overall convergence speed.
\end{enumerate}
This analysis confirms that our impurity scheduler is a highly effective component, but its optimal configuration requires careful tuning based on the specific dimensional and structural characteristics of the benchmark.

\end{document}